\documentclass[12pt,onecolumn,letterpaper]{article}
\usepackage{times}
\usepackage{graphicx}
\usepackage{amsmath}
\usepackage{amssymb}
\usepackage{siunitx}
\usepackage{caption}
\usepackage{bm}
\usepackage{mathtools}
\usepackage{extdash}
\usepackage{url}

\usepackage{booktabs}
\usepackage{xcolor}
\usepackage{colortbl}

\usepackage[
backend=biber,
style=numeric,
sorting=none
]{biblatex}
\usepackage{multirow}
\newcommand{\STAB}[1]{\begin{tabular}{@{}c@{}}#1\end{tabular}}

\newcommand{\ra}[1]{\renewcommand{\arraystretch}{#1}}

\title{Scalable 3D Reconstruction From Single Particle X-Ray Diffraction Images Based on Online Machine Learning}

\author
{Jay Shenoy$^{1}$, Axel Levy$^{2, 3}$,  Frédéric Poitevin$^{3}$, Gordon Wetzstein$^{2\ast}$}

\date{}

\begin{document} 
\sisetup{mode=text}   

\pagestyle{plain}
\thispagestyle{plain}
\baselineskip15pt

\maketitle 

\begin{abstract}
X-ray free-electron lasers (XFELs) offer unique capabilities for measuring the structure and dynamics of biomolecules, helping us understand the basic building blocks of life. Notably, high-repetition-rate XFELs enable single particle imaging (X-ray SPI) where individual, weakly scattering biomolecules are imaged under near-physiological conditions with the opportunity to access fleeting states that cannot be captured in cryogenic or crystallized conditions. Existing X-ray SPI reconstruction algorithms, which estimate the unknown orientation of a particle in each captured image as well as its shared 3D structure, are inadequate in handling the massive datasets generated by these emerging XFELs. Here, we introduce X-RAI, an online reconstruction framework that estimates the structure of a 3D macromolecule from large X-ray SPI datasets. X-RAI consists of a convolutional encoder, which amortizes pose estimation over large datasets, as well as a physics-based decoder, which employs an implicit neural representation to enable high-quality 3D reconstruction in an end-to-end, self-supervised manner. We demonstrate that X-RAI achieves state-of-the-art performance for small-scale datasets in simulation and challenging experimental settings and demonstrate its unprecedented ability to process large datasets containing millions of diffraction images in an online fashion. These abilities signify a paradigm shift in X-ray SPI towards real-time capture and reconstruction.
\end{abstract}

\footnotetext[1]{Department of Computer Science, Stanford University, Stanford, CA}
\footnotetext[2]{Department of Electrical Engineering, Stanford University, Stanford, CA}
\footnotetext[3]{LCLS, SLAC National Accelerator Laboratory, Menlo Park, CA}
\let\thefootnote\relax\footnotetext{*Corresponding author. E-mail: gordon.wetzstein@stanford.edu}

\section*{Introduction}
\label{sec:main}
Understanding the structure of proteins, viruses, and other biomolecules reveals insight into their function, helping scientists understand the building blocks of life and design new therapies~\cite{congreve2005keynote,jorgensen2004many, frauenfelder1991energy,henzler2007dynamic}. X-ray free-electron lasers (XFELs) offer the unprecedented opportunity of observing these structures under near-physiological conditions at room temperature via single particle imaging (SPI)~\cite{starodub2012single,aquila2015linac}. The unique capabilities of X-ray SPI promise to overcome the limitations of established techniques such as X-ray crystallography~\cite{smyth2000x,shi2014glimpse} and cryogenic electron microscopy (cryo-EM)~\cite{bai2015cryo,cheng2015primer,doerr2016single,cheng2018single,nakane2020single}, where the processes of crystallization and freezing alter or restrict the dynamics of biomolecules~\cite{ourmazd2019cryo,acharya2005advantages,depristo2004heterogeneity}.

\begin{figure}[t]
	\centering
	\includegraphics[width=\columnwidth]{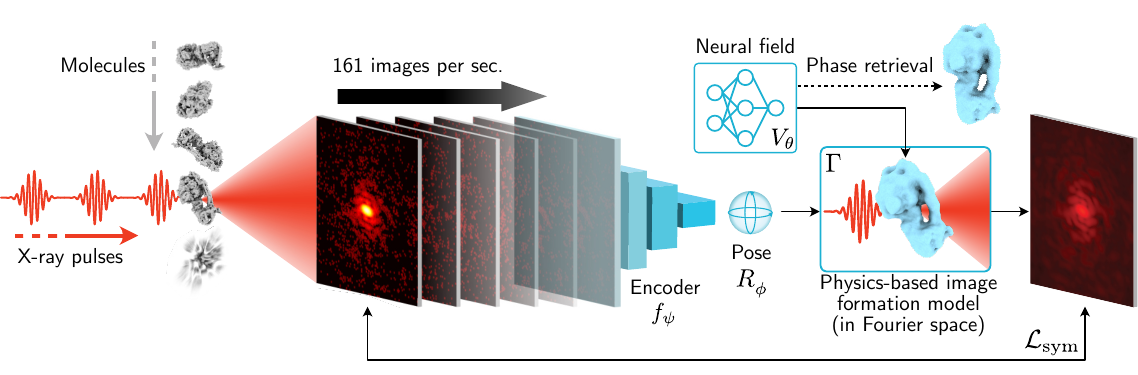}
	\caption{\textbf{Illustration of XFEL SPI image acquisition process and subsequent online processing by our algorithm.} A femtosecond-resolution X-ray pulse intersects a single, hydrated molecule, creating a diffraction image. This process is repeated at high rates to collect millions of diffraction images, each observing the molecule with an unknown orientation, or pose. Our algorithm, X-RAI, employs a CNN-based encoder $f_{\psi}$ to efficiently estimate the pose $R_\phi$ of the molecule in each image. A physically accurate decoder $\Gamma$ (in Fourier space, shown here in real space) produces a noise-free estimate of the diffraction pattern using the molecule's 3D structure, represented as a neural field $V_\theta$. The symmetric loss $\mathcal{L}_{\textrm{sym}}$ is applied to optimize the parameters $\psi$ and $\theta$ in an online fashion using self-supervision. At any point during the experiment, the volume $ V_\theta $ can be phased to obtain an estimate of the electron density.
	}
	\label{fig:pipeline}
\end{figure}

To overcome the weak signal present in single-particle snapshots and properly sample the conformational landscape of a biomolecule, nascent high-repetition-rate XFELs such as the European XFEL~\cite{sobolev2020megahertz} and the Linac Coherent Light Source (LCLS) II~\cite{schoenlein2017linac} at SLAC National Accelerator Laboratory will soon capture datasets containing a few million to hundreds of millions of diffraction images per experiment to enable X-ray SPI at scale. Existing reconstruction algorithms for X-ray SPI \cite{ayyer2016dragonfly, hosseinizadeh2017conformational, donatelli2017reconstruction, von2018structure} are unable to efficiently process datasets of this size because they conduct an exhaustive search to estimate the unknown orientations, or poses, of the particle within each diffraction image. This search strategy simultaneously and independently maximizes the likelihood of all the images and does not scale to large datasets.

Here, we introduce X-RAI, a scalable computational approach to 3D reconstruction from X-ray SPI diffraction images enabled by recent advancements in artificial intelligence. X-RAI consists of a convolutional encoder that directly maps diffraction images to predicted poses, thereby amortizing pose estimation over the entire dataset and averting the expensive pose search procedure found in prior work. The predicted poses are then mapped back to reconstructions of the input images using a physics-based decoder that employs a neural implicit representation~\cite{sitzmann2020implicit} to store the reconstructed 3D intensity. This decoder simulates the formation of the diffraction pattern on the detector arising from the interaction of an incoming X-ray pulse with the electron density of a randomly oriented copy of the particle, following the physical principles of X-ray diffraction theory \cite{ewald1969introduction} to ensure that no information is hallucinated. The reconstruction algorithm processes diffraction images in sequential batches to compute updates to the encoder and volume representation in a self-supervised, end-to-end fashion. Each batch therefore contributes an incremental update to the encoder--decoder model, and viewed this way, the problem of molecular reconstruction becomes one of incremental online learning \cite{wang2023comprehensive}, wherein the model learns from data acquired sequentially rather than simultaneously.

In the online setting, each image in the dataset is given to X-RAI only once while other algorithms iterate over all the images throughout several epochs. Remarkably, X-RAI is able to perform reconstruction on synthetic datasets containing millions of images in a completely online fashion, demonstrating the feasibility of live reconstruction within an actual experiment. We also validate X-RAI's accuracy in the non-incremental, offline setting by demonstrating reconstruction on smaller datasets with equivalent or better quality than prior techniques. Further, we test our method on the largest available experimental dataset of a biological sample collected at an XFEL, achieving a resolution near the detector limit. Our results demonstrate the efficacy of an amortized end-to-end approach for both online and offline reconstruction, offering a novel computational paradigm that promises to unlock the potential of data-heavy experiments that are emerging in X-ray SPI.

\section*{Results}
\label{sec:results}
\subsection*{Image Formation and Reconstruction}
\label{sec:image_formation_reconstruction}

In X-ray SPI, when a laser pulse intersects a molecule, the light diffracts to form an intensity image on the detector. As described by diffraction theory~\cite{ewald1969introduction}, this intensity pattern samples the Fourier transform of the electron density along a geometric construct known as the Ewald sphere~\cite{ewald1921berechnung}. The resulting image on the detector represents the squared amplitude of the Fourier transform, losing phase information. We assume that the particle in each image is oriented randomly with respect to the laser and that the light incident on the detector is subject to Poisson-distributed photon noise. As such, the intensity measured at pixel $ (u, v) $ of the diffraction image can be written as
\begin{equation}
    I(u,v;V,R_\phi)\sim \text{Pois}\Bigl[V\bigl(R_\phi\cdot E(u,v)\bigr)\Bigr],
    \label{eq:forward-model-1}
\end{equation}

where $V:\mathbb{R}^3\to\mathbb{R}_+$ represents the intensity volume, i.e., the squared amplitude of the Fourier transform of the 3D electron density volume, $R_\phi\in\text{SO(3)}\subset\mathbb{R}^{3\times 3}$ is the orientation of the particle, $\text{Pois}[\eta]$ is a homogeneous Poisson process with rate $\eta$, and $E:\mathbb{R}^2\to\mathbb{R}^3$ is the Ewald sphere.
A detector defines a set of $N$ pixels $\{(u_k, v_k)\}_{k\in\{1,\ldots,N\}}$.

To estimate the 3D electron density from a set of diffraction patterns in an online fashion, X-RAI uses an encoder--decoder network architecture (Figure~\ref{fig:pipeline}). As images are collected, our method optimizes a pair of neural networks: the convolutional encoder $f_\psi : \mathbb{R}^{N_u \times N_v} \to \mathbb{R}^{6}, I \mapsto f_\psi \left( I \right)$, mapping images with a resolution of $N_u \times N_v$ pixels to poses defined by six scalars, and the implicit neural volume representation $V_\theta : \mathbb{R}^3 \to \mathbb{R}_+, \left( k_x, k_y, k_z \right) \mapsto V_\theta \left( k_x, k_y, k_z  \right)$, mapping coordinates in Fourier space to intensities (see Methods for additional details). We first apply a variance-stabilizing transformation, $Y=\sqrt{I}$, to the diffraction patterns to make the noise model approximately Gaussian~\cite{bartlett1936square}.

This transformation aligns the noise model of the measurements with that expected by the encoder and makes the L2 norm in image space a good approximation of the negative log-probability of the noise. Instead of independently optimizing the pose of each diffraction image, X-RAI uses the same convolutional encoder for all the images and effectively \textit{learns} to estimate poses on-the-fly. Based on Eq.~\eqref{eq:forward-model-1}, a physics-based decoder $\Gamma$ reconstructs noise-free estimates of the input diffraction images using the known geometry of the detector, the predicted orientations $R_\phi$, and an implicit neural representation of the intensity volume $V_\theta$, a process that can be formally written as:
\begin{equation}
    \Gamma:V_\theta,R_\phi\mapsto\{V_\theta\bigl(R_\phi\cdot E(u_k,v_k)\bigr)\}_{k\in\{1,\ldots,N\}}.
\end{equation}
Unlike a voxel-based representation ($\in\mathbb{R}^{N\times N\times N}$), the implicit neural representation maps 3D coordinates to intensities and can be queried continuously in Fourier space without the need to explicitly define an interpolation scheme. Furthermore, in a scenario where view directions are optimized jointly with the 3D density, implicit representations have been shown to be more adapted to gradient-based optimization than voxel arrays as they naturally learn low-frequency details first, thereby avoiding local minima~\cite{levy2023melon}.

In the online setup, collected images are sequentially processed in small batches. X-RAI successively updates the parameters of the neural networks, $\psi$ and $\theta$, so as to maximally increase the likelihood of the observed batch $\mathcal{B}_t$. Specifically, the update rule takes a step opposed to the gradient of
\begin{equation}
\mathcal{L}_\text{sym}(\psi,\theta;t) = \sum_{I\in\mathcal{B}_t}\text{min}\Bigl\{\Vert Y-\Gamma(V_{\theta},f_{\psi}(Y))\Vert_2^2, Y\in \mathcal{O}(\sqrt{I})\Bigr\}
\label{eq:sym_loss}
\end{equation}
with respect to $\psi$ and $\theta$.

Due to quasisymmetries in the ground truth 3D density, diffraction images that appear similar but are associated with distant poses tend to be mapped to the same pose by the encoder, causing spurious symmetries appear in the reconstructed volume when using a simple L2 loss. To avoid this behavior, two variants of each diffraction image, the original image and a flipped version, are run through the pipeline in parallel, as indicated by the $ \mathcal{O}(\sqrt{I}) $ term in equation~$ \ref{eq:sym_loss} $. $\mathcal{O}(Y)$ corresponds to the pair $(\{Y(u_k,v_k)\}, \{Y(v_k,u_k)\})$, and gradients are only computed for the run that produces the lower reconstruction error. Our particular choice of $\mathcal{O}(Y)$ ensures that X-RAI converges even in the presence of quasi-planar symmetries. In the online learning setup, images can be processed on-the-fly at a speed of 161 images per second on a single GPU. At any point during the experiment, the volume $ V_\theta $ can be phased to obtain an estimate of the electron density.

\subsection*{Comparison to State-of-the-Art Offline Reconstruction}
\label{sec:offline_comparison}

\begin{figure*}[t]
	\centering
	\includegraphics[width=\columnwidth]{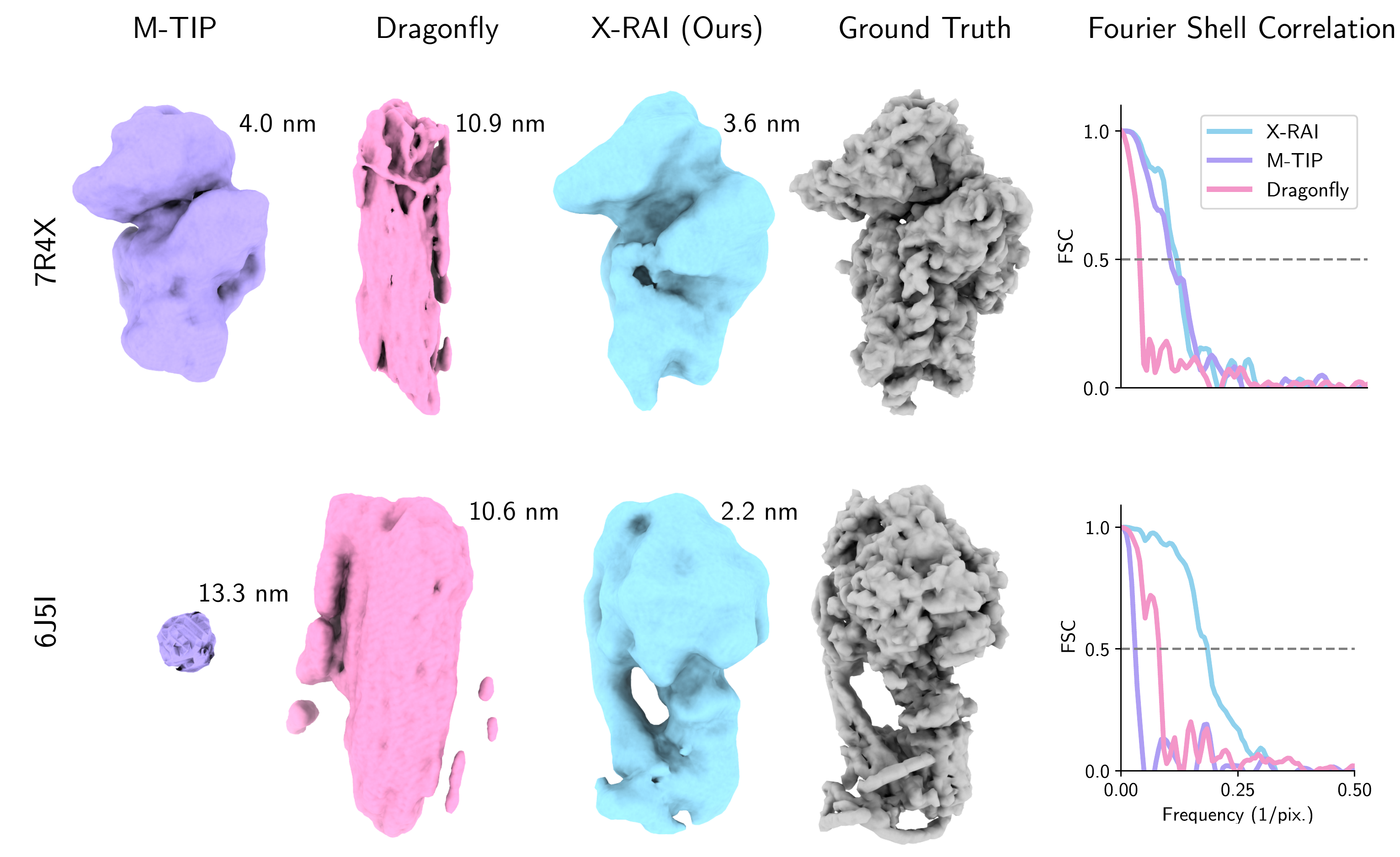}
	\caption{
            \textbf{Comparison of X-RAI with M-TIP \cite{donatelli2017reconstruction} and Dragonfly \cite{ayyer2016dragonfly} on reconstructing datasets with 50,000 images in an offline setting.} Each row corresponds to a separate dataset simulated using Skopi \cite{peck2022skopi} with the protein data bank (PDB) code specified in the leftmost column. For both datasets, X-RAI is able to reconstruct each particle to within 4~nanometers of resolution, exceeding the quality of the other two methods. For the protein 6J5I, M-TIP fails to reconstruct the intensity volume, resulting in the degenerate density volume shown. The reconstructions shown for Dragonfly are scaled down in size to fit into the figure.
        }
	\label{fig:offline_comparison}
\end{figure*}

We first validate our method on the problem of offline reconstruction, demonstrating reconstruction quality that matches or exceeds that of prior work. X-RAI can be trained offline on smaller datasets by conducting several epochs over the data until convergence. To compare our method’s performance against that of M-TIP \cite{donatelli2017reconstruction} and Dragonfly \cite{ayyer2016dragonfly}, we simulate several synthetic datasets containing 50,000 images each using the software package Skopi \cite{peck2022skopi}. Two of these, the 40S ribosomal subunit (PDB: 7R4X) \cite{pellegrino2023cryo} and the ATP synthase molecule (PDB: 6J5I) \cite{gu2019cryo}, are shown in Figure~\ref{fig:offline_comparison}. X-RAI is able to reconstruct both particles to sub-4 nanometer resolution, surpassing the quality of the other two methods. All volumes shown in the figure represent the reconstructed densities after running the algorithms to convergence. Although we run M-TIP ten times independently and select the best reconstruction, it fails to reconstruct the ATP synthase protein (6J5I) during all the attempts, resulting in the degenerate electron density volume shown. Our method’s superior performance arises from the accuracy of the end-to-end encoder--decoder approach at pose estimation, which can be seen in Table \ref{table:offline_pose_accuracy_comparison}. X-RAI estimates molecular orientations to within 9 degrees of accuracy, whereas M-TIP and Dragonfly predict poses that have at least 11 degrees of error in the best case and random orientations in the extreme.

\subsection*{Online Single Particle Reconstruction}
\label{sec:online_reconstruction}

\begin{figure*}[t]
	\centering
	\includegraphics[width=\columnwidth]{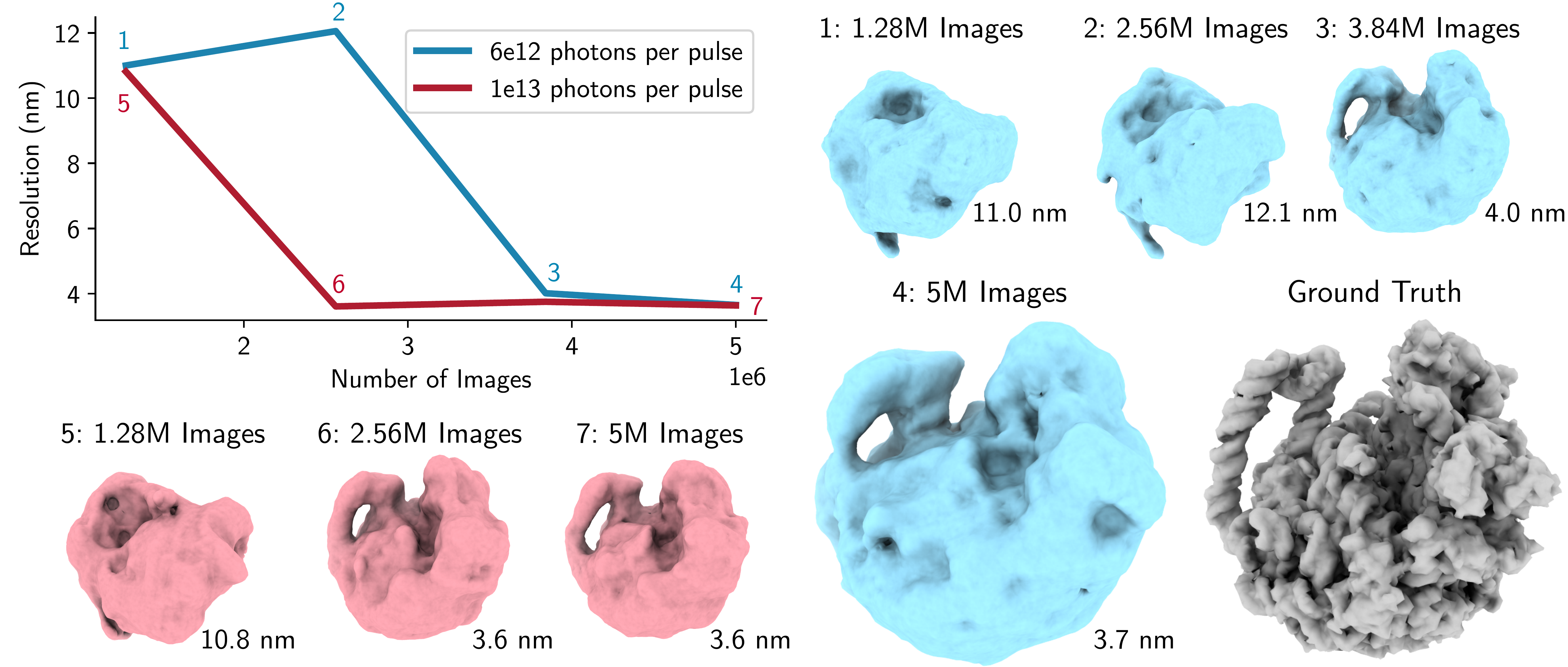}
	\caption{
            \textbf{Online reconstruction of the 50S ribosomal subunit (PDB: 5O60) \cite{hentschel2017complete}.} Remarkably, X-RAI is able to reconstruct the protein even when acquiring the data sequentially in batches of $ 64 $ images. Two synthetic datasets with varying levels of beam fluence (photons per pulse) are shown in order to illustrate the effect of signal level on convergence speed. Our method converges to a resolution of about $ 3.6 $ nm after processing 2.56 million images when handling images with $ 10^{13} $ photons per pulse. On the other hand, the reconstruction only converges after around 3.84 million images have been processed for the dataset with a lower fluence of $ 6 \times 10^{12} $ photons per pulse, demonstrating the importance of large datasets in the low signal regime.
	}
	\label{fig:resolution_test}
\end{figure*}

For large datasets containing millions of images, X-RAI can be run online by processing data in small batches, each of which contributes a gradient update to the model. This amounts to running stochastic gradient descent over the entire dataset for one epoch, and X-RAI is able to process data at a speed of 161 images per second on a single GPU. Figure \ref{fig:resolution_test} illustrates how reconstruction quality improves as X-RAI acquires synthetic data of the 50S ribosomal subunit (PDB: 5O60) \cite{hentschel2017complete} in an online fashion for two levels of beam fluence. For the higher fluence level corresponding to $ 10^{13} $ photons per pulse, the diffraction images have a higher level of signal and only require around two million images to converge to a resolution of 3.6 nanometers, compared to around three million images for a fluence of $ 6 \times 10^{12} $ photons per pulse. Moreover, reconstruction at the lower fluence level demonstrates the importance of large datasets in obtaining sharper density volumes. As shown in Table~\ref{table:offline_pose_accuracy_comparison}, the resolution improves from 5.7 nm for 50 thousand images (run offline for 100 epochs) to 3.7 nm for 5 million images (processed online). 

Online reconstruction is fundamentally more difficult than its offline counterpart because when reconstruction is run for more than one epoch, the encoder can potentially memorize the mapping of input images to poses (given a sufficiently small dataset) rather than learning the true function that relates these two quantities.

Our method’s success at online reconstruction shows that the model determines the true association between diffraction patterns and orientations without relying on memorization.

Table~\ref{table:offline_pose_accuracy_comparison} contains quantitative results for nine synthetic datasets consisting of three proteins simulated at three dataset sizes. Remarkably, X-RAI is able to process all the datasets containing 5 million images in an online manner. Each method is run for a maximum of 48 hours and terminated if it exceeds the time or memory budget. M-TIP is unable to process datasets with 500 thousand or 5 million images, and Dragonfly exceeds the time limit for 5 million images, as signified by the empty rows in the table. Visual reconstruction results for all the synthetic datasets can be found in the supplement.

\begin{table}[ht]
\footnotesize
\centering
\ra{1.4}
 \begin{tabular}{@{} l l l c c c c c @{}} 
 \toprule
  & \# Im. & Method & Time (h) & Resolution (nm) $ \downarrow $ & Median Pose Error (\textdegree) $ \downarrow $ & Mean Pose Error (\textdegree) $ \downarrow $ \\ [0.5ex] 
 \midrule
 \multirow{9}{*}{\STAB{\rotatebox[origin=c]{90}{7R4X (fluence: $ 10^{14} $)}}} & 50k & M-TIP & 6:05 & 4.0 & 11.5 & 32.8 \\
  & & Dragonfly & 1:20 & 10.9 & 81.5 & 83.1 \\
  & & X-RAI & 8:36 & \textbf{3.6} & \textbf{4.7} & \textbf{8.1} \\
 \cmidrule{3-7}
  & 500k & M-TIP & ---- & ---- & ---- & ---- \\
  & & Dragonfly & 22:19 & 9.8 & 55.2 & 67.6 \\
  & & X-RAI & 8:38 & \textbf{4.1} & \textbf{2.1} & \textbf{4.1} \\
 \cmidrule{3-7}
  & 5M & M-TIP & ---- & ---- & ---- & ---- \\
  & & Dragonfly & ---- & ---- & ---- & ---- \\
  & & X-RAI & 8:41 & \textbf{4.0} & \textbf{3.3} & \textbf{7.2} \\
 \cmidrule{2-7}
 \multirow{9}{*}{\STAB{\rotatebox[origin=c]{90}{6J5I (fluence: $ 10^{13} $)}}} & 50k & M-TIP & 6:30 & 13.3 & 88.9 & 89.1 \\
  & & Dragonfly & 3:10 & 10.6 & 67.8 & 70.1 \\
  & & X-RAI & 8:36 & \textbf{2.2} & \textbf{1.3} & \textbf{1.6} \\
  \cmidrule{3-7}
  & 500k & M-TIP & ---- & ---- & ---- & ---- \\
  & & Dragonfly & 34:28 & 11.7 & 84.4 & 85.6 \\
  & & X-RAI & 8:38 & \textbf{2.1} & \textbf{1.5} & \textbf{2.0} \\
  \cmidrule{3-7}
  & 5M & M-TIP & ---- & ---- & ---- & ---- \\
  & & Dragonfly & ---- & ---- & ---- & ---- \\
  & & X-RAI & 8:37 & \textbf{2.0} & \textbf{1.6} & \textbf{1.9} \\
  \cmidrule{2-7}
 \multirow{9}{*}{\STAB{\rotatebox[origin=c]{90}{5O60 (fluence: $ 6 \times 10^{12} $)}}} & 50k & M-TIP & 6:52 & 13.3 & 61.2 & 64.8 \\
  & & Dragonfly & 1:17 & 13.2 & 83.9 & 84.8 \\
  & & X-RAI & 8:52 & \textbf{5.7} & \textbf{1.4} & \textbf{1.5} \\
  \cmidrule{3-7}
  & 500k & M-TIP & ---- & ---- & ---- & ---- \\
  & & Dragonfly & 13:19 & 13.8 & 84.4 & 85.1 \\
  & & X-RAI & 8:39 & \textbf{4.1} & \textbf{2.3} & \textbf{2.5} \\
  \cmidrule{3-7}
  & 5M & M-TIP & ---- & ---- & ---- & ---- \\
  & & Dragonfly & ---- & ---- & ---- & ---- \\
  & & X-RAI & 8:36 & \textbf{3.7} & \textbf{2.5} & \textbf{2.9} \\
 \bottomrule \\
 \end{tabular}
 \caption{\textbf{Quantitative comparison of X-RAI, M-TIP, and Dragonfly.} For all datasets, X-RAI achieves the best pose estimation accuracy and resolution. Notably, since we fix the number of batches processed by X-RAI (training for 100, 10, and 1 epoch(s) for 50k, 500k, and 5M images, respectively), its reconstruction time remains constant for all three dataset sizes.}
 \label{table:offline_pose_accuracy_comparison}
\end{table}

\subsection*{Reconstruction From Experimental Data}
\label{sec:experimental]}

In addition to validating X-RAI on synthetic datasets in the online and offline settings, we also test its performance on a real experimental dataset captured at an XFEL. Consisting of 14,772 diffraction images of the PR772 coliphage virus, this dataset \cite{reddy2017coherent} is the largest biological collection of its kind from an SPI experiment. Owing to the virus's size, the sample scatters hundreds of thousands of photons per image, which is significantly larger than our simulated proteins.

We reconstruct PR772 by introducing a novel loss formulation, the \textit{randomized symmetric loss}, to enforce the known icosahedral symmetry of the virus. During training, we randomly rotate the poses output by the encoder by a symmetry rotation of the icosahedron, an operation that can be expressed as a modification of the original loss function in equation \ref{eq:sym_loss}:
\begin{equation}
    \mathcal{L}_\text{sym}(\psi,\theta;t) = \sum_{i=1}^{|\mathcal{B}_t|}\text{min}\Bigl\{\Vert Y-\Gamma(V_\theta, R_{i,t} \cdot f_\psi(Y)\Vert^2, Y\in \mathcal{O}(\sqrt[10]{I_{i,t}})\Bigr\},
\end{equation}
where $R_{i,t}$ represents a randomly-selected symmetry rotation. Due to the symmetries of the virus, the encoder can only map diffraction patterns to a subset of $\text{SO}(3)$ (see supplement). In the equation above, $ \mathcal{O} $ is defined in the same manner as in the symmetric loss function (equation~\ref{eq:sym_loss}), and we apply the tenth root operator to the diffraction images in order to reduce the high image contrast caused by the abundance of photon hits at low frequencies. $ R_{i,t} $ belongs to the icosahedral rotation group, which in its orientation-preserving form consists of 60 rotation matrices that produce symmetric views when applied to the virus's density and intensity volumes. After the encoder outputs a pose estimate, the system rotates this estimate by a randomly-chosen rotation matrix from the icosahedral group, and we jointly optimize the orientation of the three axes defining this symmetry group during reconstruction. Since the rotation matrix is selected randomly for each image in the batch, icosahedral symmetry is enforced in a soft manner over all the batches.

\begin{figure*}[t]
	\centering
	\includegraphics[width=\columnwidth]{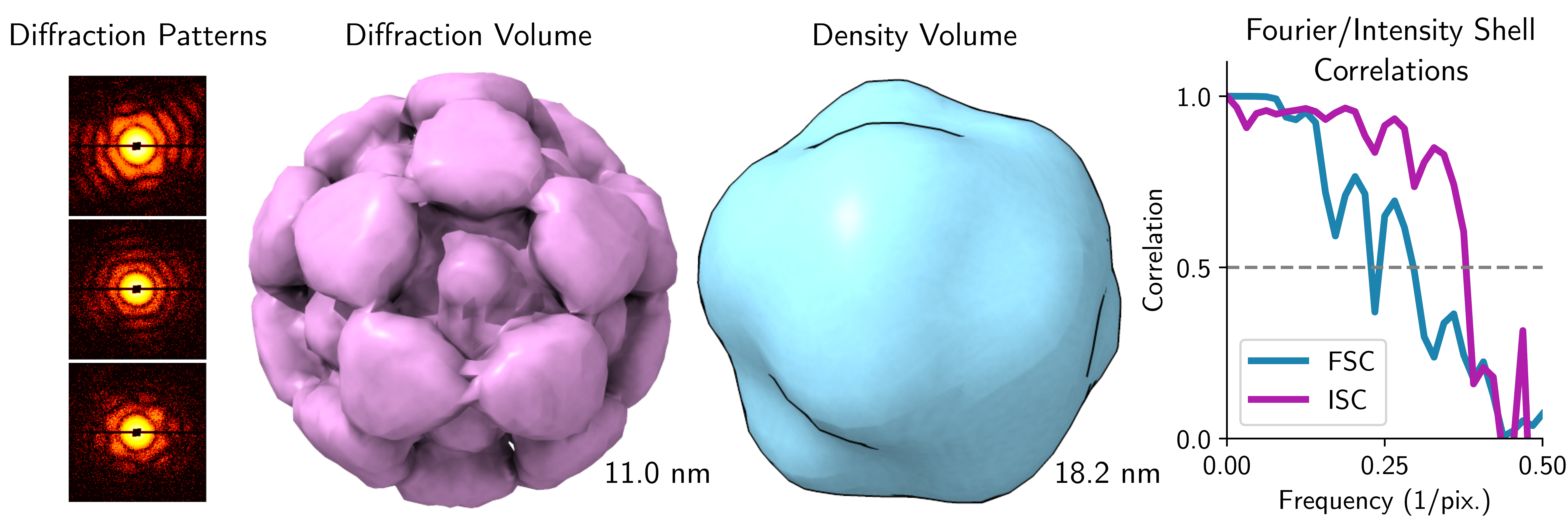}
	\caption{
            \textbf{Reconstruction of the PR772 virus from experimental data collected at an XFEL facility~\cite{reddy2017coherent}.}
            To enforce the known icosahedral symmetry of the virus, we augment the encoder's orientation estimates with the symmetry rotations of the icosahedron, effectively spreading out the pose estimates over $\text{SO}(3)$. The resulting intensity and density reconstructions are of resolution 11.0 and 18.2~nanometers, respectively. The diffraction patterns are displayed with reduced image contrast.
        }
	\label{fig:experimental_reconstruction}
\end{figure*}

This replication scheme effectively spreads out the encoder's pose estimates over $\text{SO}(3)$ and enables the intensity volume to be supervised on the entirety of $\mathbb{R}^3$, resulting in the reconstructions shown in Figure~\ref{fig:experimental_reconstruction}. Optimization is run for 100 epochs on this dataset. Comparing reconstructions on two random subsets of the data, we achieve a resolution of 11.0~nm on the intensity (diffraction) volume and a resolution of 18.2~nm on the density. The discrepancy between these two quantities likely arises from the stochasticity of the phase retrieval algorithm. By comparison, the best previously reported symmetric reconstruction reaches a resolution of 9~nm on a filtered version of the dataset, leveraging strict enforcement of the icosahedral symmetry in real space and lacking the ability to run online~\cite{hosseinizadeh2017conformational}. The absolute resolution limit of this dataset, as determined by the coordinates of the detector corner, is 8.3~nm~\cite{reddy2017coherent}.

\section*{Discussion}
\label{sec:discussion}
The method introduced in this work demonstrates the efficacy of an encoder--decoder approach leveraging a physics-based differentiable simulator for protein reconstruction in X-ray SPI. By amortizing the estimation of pose over the entire dataset, X-RAI avoids the expensive orientation search procedures found in prior work and achieves qualitative and quantitative reconstructions that match the state-of-the-art. Moreover, the end-to-end differentiability of our system allows it to conduct reconstruction in an online setting where data are acquired incrementally, an approach that demarcates X-RAI from existing methods that process datasets in batch.

With emerging XFELs poised to deliver megahertz repetition rate~\cite{sobolev2020megahertz}, scalable reconstruction algorithms are becoming increasingly important to the success of XFEL experiments, providing live feedback and informing data collection and reduction strategies. Because X-RAI relies on stochastic gradient descent, its image throughput rate depends mainly on the batch size, which can easily be increased by parallelizing over an arbitrary number of GPUs. Another consideration for real SPI experiments is that smaller molecules will scatter fewer photons and therefore produce sparser diffraction images. In this work, we primarily validate our method on synthetic and real datasets with relatively high photon counts, so evaluating and improving upon our framework in the low signal regime is a promising avenue for future research.

The ultimate opportunity of X-ray SPI lies in recovering the transient, continuous dynamics that have eluded methods such as cryo-EM and crystallography \cite{bock2022effects, ourmazd2019cryo}. Future XFEL experiments will collect large datasets that capture these short-lived states, requiring the development of efficient algorithms that can handle biomolecular dynamics \cite{shi2019evaluation,asi2022hybrid,zhuang2022unsupervised}. The reconstructions presented here are of homogeneous molecules, so further work must be done to validate our method in the heterogeneous setting, building on prior techniques that equip related network architectures with mechanisms that handle conformational diversity \cite{zhong2021cryodrgn, levy2022cryofire}. Moreover, the vast amounts of data already being collected at the European XFEL and expected soon at LCLS-II will benefit from on-the-fly processing in several ways. First, the high repetition rate makes storage of raw data quite challenging, if not impossible. While generic data reduction strategies will help alleviate the pressure on data systems, adapting online analysis to particular experiments could ultimately prove to be a more efficient strategy. In addition, real-time analysis provides direct quality metrics of the experiment status, enabling faster decision-making and thus optimal use of beamtime. Overall, X-RAI represents a powerful approach to SPI reconstruction that compares favorably to existing techniques and offers an online framework that promises to scale to the large datasets currently in sight at XFEL facilities.


\clearpage

\section*{Methods}
\label{sec:methods}
\paragraph{Image formation model.} 

In X-Ray SPI, X-ray pulses interact with the 3D electron density of individual molecules. We assume that all the molecules share the same density function and refer to $R_\phi\in\text{SO}(3)\subset\mathbb{R}^{3\times3}$ as the orientation of the molecule with respect to a canonical orientation when hit by a pulse of light. After being diffracted by the particle, the pulse of light (with wavelength $\lambda$) reaches a detector placed at a distance $d$. The intensity $I$ of the light on the detector varies and is subject to Poisson-distributed photon noise. Following diffraction theory~\cite{ewald1969introduction}, the diffraction signal can be modeled as a space-dependent random variable following a Poisson distribution
\begin{equation}
    I(u,v;V,R_\phi) \sim \text{Pois}\Bigl[|V(R_\phi\cdot E(u,v))|^2\Bigr]
\end{equation}
where $V:\mathbb{R}^3\to\mathbb{R}_+$ is the square amplitude of the Fourier transform of the electron density, $\text{Pois}[\eta]$ is a homogeneous Poisson process with rate $\eta$, and $E:\mathbb{R}^2\to\mathbb{R}^3$ is the Ewald sphere:
\begin{equation}
    E(u,v)=\begin{pmatrix}
\frac{1-\alpha(u,v)}{\alpha(u,v)\lambda}\\
\frac{u}{d}\cdot\frac{1}{\alpha(u,v)\lambda}\\
\frac{v}{d}\cdot\frac{1}{\alpha(u,v)\lambda}
\end{pmatrix},\quad\text{where}\quad\alpha(u,v)=\sqrt{1+\frac{u^2}{d^2}+\frac{v^2}{d^2}}.
\end{equation}
A detector defines a set of $N$ 2D coordinates $\{u_k,v_k\}_{k\in\{1,\ldots,N\}}$. A diffraction image $\Gamma(V,R_\phi)$ can therefore be seen as a set of $N$ measurements
\begin{equation}
    I=\{I(u_k,v_k;V,R_\phi)\}_{k\in\{1,\ldots,N\}}.
\end{equation}

\paragraph{Implicit neural representation.} Instead of using an explicit voxel array to approximate the function $V$, we parameterize $V_\theta$ with a neural network. As shown in~\cite{levy2023melon} and in the supplement, implicit representations allow the orientations to converge more robustly than do explicit representations. The architecture of our neural network is based on the FourierNet introduced in~\cite{levy2022cryoai}. It is built with two Sinusoidal Implicit Representation Networks (SIRENs~\cite{sitzmann2020implicit}) with 256 hidden features, mapping $\mathbb{R}^3$ to $\mathbb{R}^2$. They contain 2 and 3 hidden layers, respectively. We take the exponential of the output of the \textit{small} network and multiply it element-wise with the output of the \textit{large} network. The total number of parameters is 332,036 (vs. $128^3$ = 2,097,152 for an explicit representation).

\paragraph{Encoder.} X-RAI uses a convolutional neural network (CNN) to map diffraction images to poses. The architecture of the encoder has three layers. First, the diffraction image is fed to a bank of Gaussian low-pass filters. Then, the filtered images are stacked channel-wise and fed into a CNN whose architecture is inspired by the first layers of VGG16~\cite{simonyan2014very}, known to perform well on image classification tasks. Finally, the feature vector output by the CNN becomes the input of a
fully-connected neural network that outputs a 6-dimensional vector in $S^2\times S^2$~\cite{zhou2019continuity} (two vectors on the unitary sphere in $\mathbb{R}^3$) and converted into a matrix $R_\phi \in \mathbb{R}^{3\times3}$ using the PyTorch3D library~\cite{ravi2020accelerating}. The total number of parameters in the encoder is 4,760,006.

\paragraph{Optimization.} All the parameters of the model ($\psi,\theta$) are optimized by stochastic gradient-descent, using the Adam~\cite{kingma2014adam} optimizer with a learning rate of $10^{-4}$. The batch size is set to 64.

\paragraph{Hardware.} All our experiments are run on a single Tesla A100 GPU.

\paragraph{Visualization.} Molecular graphics and analyses are performed with UCSF ChimeraX~\cite{pettersen2021ucsf}, developed by the Resource for Biocomputing, Visualization, and Informatics at the University of California, San Francisco, with support from National Institutes of Health R01-GM129325 and the Office of Cyber Infrastructure and Computational Biology, National Institute of Allergy and Infectious Diseases. In all the figures, we show a single isosurface of the electron density.

\printbibliography

\section*{Acknowledgments}

J.S. is supported by the National Science Foundation Graduate Research Fellowship under Grant No. DGE-2146755 and the SLAC LDRD project ``AtomicSPI: Learning atomic scale biomolecular dynamics from single-particle imaging data'' (PI: F.P.). This work was supported by the U.S. Department of Energy, under DOE Contract No. DE-AC02-76SF00515. We acknowledge the use of the computational resources at the SLAC Shared Scientific Data Facility (S3DF). G.W. is supported by the ARO (PECASE Award W911NF-19-1-0120).

\end{document}


\sisetup{mode=text}   

\pagestyle{plain}
\thispagestyle{plain}
\baselineskip15pt

\maketitle 
\noindent \textbf{This PDF file includes:}\\
\indent Supplementary Note 1: Ablation study\\
\indent Supplementary Note 2: Replication scheme on PR772\\
\indent Supplementary Note 3: Datasets\\
\indent Supplementary Note 4: Qualitative reconstruction results on all synthetic datasets\\
\indent Supplementary References\\

\footnotetext[1]{Department of Computer Science, Stanford University, Stanford, CA}
\footnotetext[2]{Department of Electrical Engineering, Stanford University, Stanford, CA}
\footnotetext[3]{LCLS, SLAC National Accelerator Laboratory, Menlo Park, CA}
\let\thefootnote\relax\footnotetext{*Corresponding author. E-mail: gordon.wetzstein@stanford.edu}

\newpage

\section*{Supplementary Note 1: Ablation study} 

\begin{figure}[h!]
	\centering
	\includegraphics[width=\columnwidth]{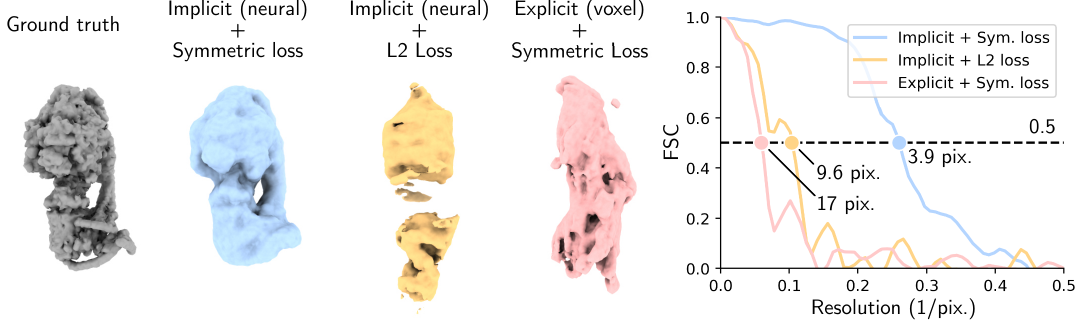}
	\caption{Ablation study on the symmetric loss function and on the implicit neural representation. (Left) Reconstructed densities with X-RAI (implicit representation + symmetric loss) and comparison with modified versions of the framework (L2 loss instead of symmetric loss and voxel-based representation instead of neural representation). (Right) Associated Fourier Shell Correlations.
	}
	\label{fig:supp:ablation-study}
\end{figure}

We conduct an ablation study and report the results in Fig.~\ref{fig:supp:ablation-study}. Specifically, we analyze the importance of the implicit neural representation ($V_\theta:\mathbb{R}^3\to\mathbb{R}_+$) by replacing it with an explicit representation ($V_\theta\in\mathbb{R}_+^{N\times N\times N}$) (i.e., a voxel grid) and that of the symmetric loss function by replacing it with a naive L2 loss
\begin{equation}
    \mathcal{L}(\psi,\theta;t)=\sum_{I\in\mathcal{B}_t}\Vert Y - \Gamma(V_\theta,f_\psi(Y))\Vert_2^2,
\end{equation}
where $Y=\sqrt{I}$. Our results show that both components are necessary for converging to the correct density. 
\vspace{48pt}

\newpage

\section*{Supplementary Note 2: Replication scheme on PR772} 

\begin{figure}[h!]
	\centering
	\includegraphics[width=\columnwidth]{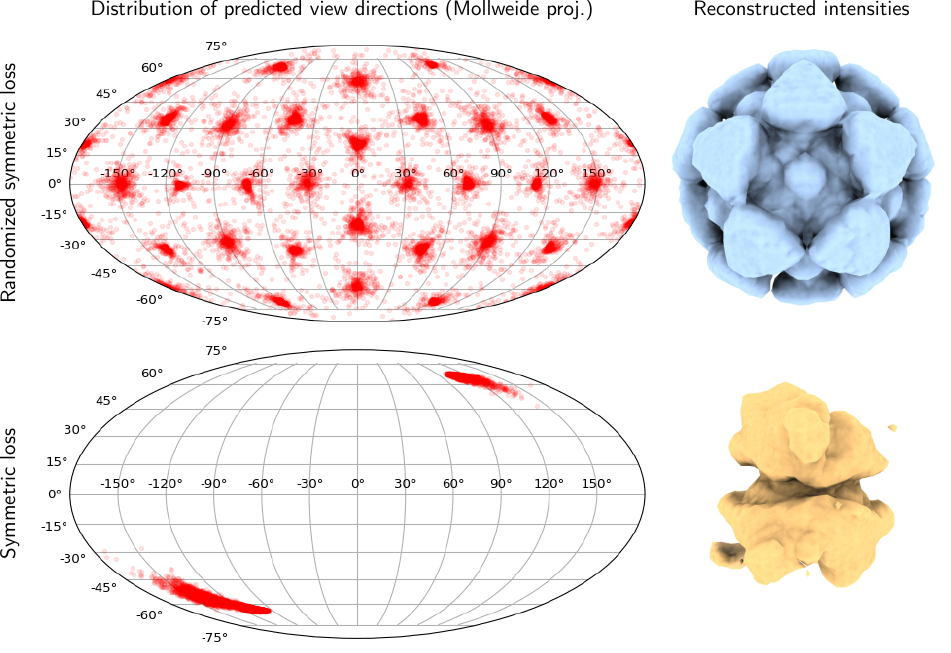}
	\caption{Importance of the randomized symmetric loss when performing reconstruction on the PR772 dataset~\cite{reddy2017coherent}. Due to the icoshedral symmetry of the virus, diffraction patterns associated with different poses can be indistinguishable (except for the noise, which should not be a source of information). Therefore, the encoder maps images to a subset of $\text{SO}(3)$ (bottom left). Without the replication scheme, the intensity volume, supervised on a subset of $\mathbb{R}^3$, is inaccurate (bottom right). With the replication scheme, the encoder spreads out predicted poses over $\text{SO}(3)$ (upper left), and the intensity volume is supervised everywhere (upper right).
	}
	\label{fig:supp:replication-loss}
\end{figure}

We illustrate the importance of the randomized symmetric loss in Fig.~\ref{fig:supp:replication-loss}. With the randomized symmetric loss, predicted poses are randomly rotated by a symmetry rotation of the icosahedron. This effectively enables the intensity volume to be supervised everywhere in $\mathbb{R}^3$, although the encoder only maps images to a subset of $\text{SO}(3)$.
\vspace{48pt}

\newpage

\section*{Supplementary Note 3: Datasets}

We simulate all synthetic datasets using the Skopi software package~\cite{peck2022skopi}. All the simulation parameters, as well as the statistics of the synthetic and experimental datasets, are detailed in Table~\ref{table:dataset_information}.

\begin{table}[h!]
\scriptsize
\centering
\ra{1.4}
 \begin{tabular}{@{} l l l l l l l @{}} 
 \toprule
 ID & Type & \# Images & Res. (pix.) & Beam Fluence & Photons per Frame & Corner Res. \\ [0.5ex] 
 \midrule

 7R4X \cite{pellegrino2023cryo} & Synthetic & 50K, 500K, 5M & 128 $ \times $ 128 & $ 10^{13} $ &
 $1.1\times10^4$
 & 8.0~\AA \\
 6J5I \cite{gu2019cryo} & Synthetic & 50K, 500K, 5M & 128 $ \times $ 128 & $ 10^{14} $ &
 $3.3\times10^4$
 & 8.0~\AA \\
 5O60 \cite{hentschel2017complete} & Synthetic & 5M & 128 $ \times $ 128 & $ 6 \times 10^{12} $ &
 $8.7\times10^3$
 & 8.0~\AA \\
 5O60 & Synthetic & 5M & 128 $ \times $ 128 & $ 10^{13} $ &
 $1.5\times10^4$
 & 8.0~\AA \\
 PR772 \cite{reddy2017coherent} & Experimental & 14,772 & 260 $ \times $ 257 & $ 10^{13} $ &
 $4.0\times10^5$
 & 83~\AA \\
 \bottomrule \\
 \end{tabular}
 \caption{Overview of synthetic and experimental datasets used for reconstruction. For all simulations, we use a photon energy of 4.6 keV, a beam radius of 0.5 microns, and a detector with a side length of 0.1 meters that is placed 0.2 meters away from the sample. For the experimental PR772 dataset, we first crop the images to resolution $ 256 \times 256 $ and then downsample them to $ 128 \times 128 $ before running reconstruction. Beam fluence is reported in number of photons.}
 \label{table:dataset_information}
\end{table}

\newpage
\section*{Supplementary Note 4: Qualitative reconstruction results on all synthetic datasets}

\begin{figure}[ht]
	\centering
	\includegraphics[width=\columnwidth]{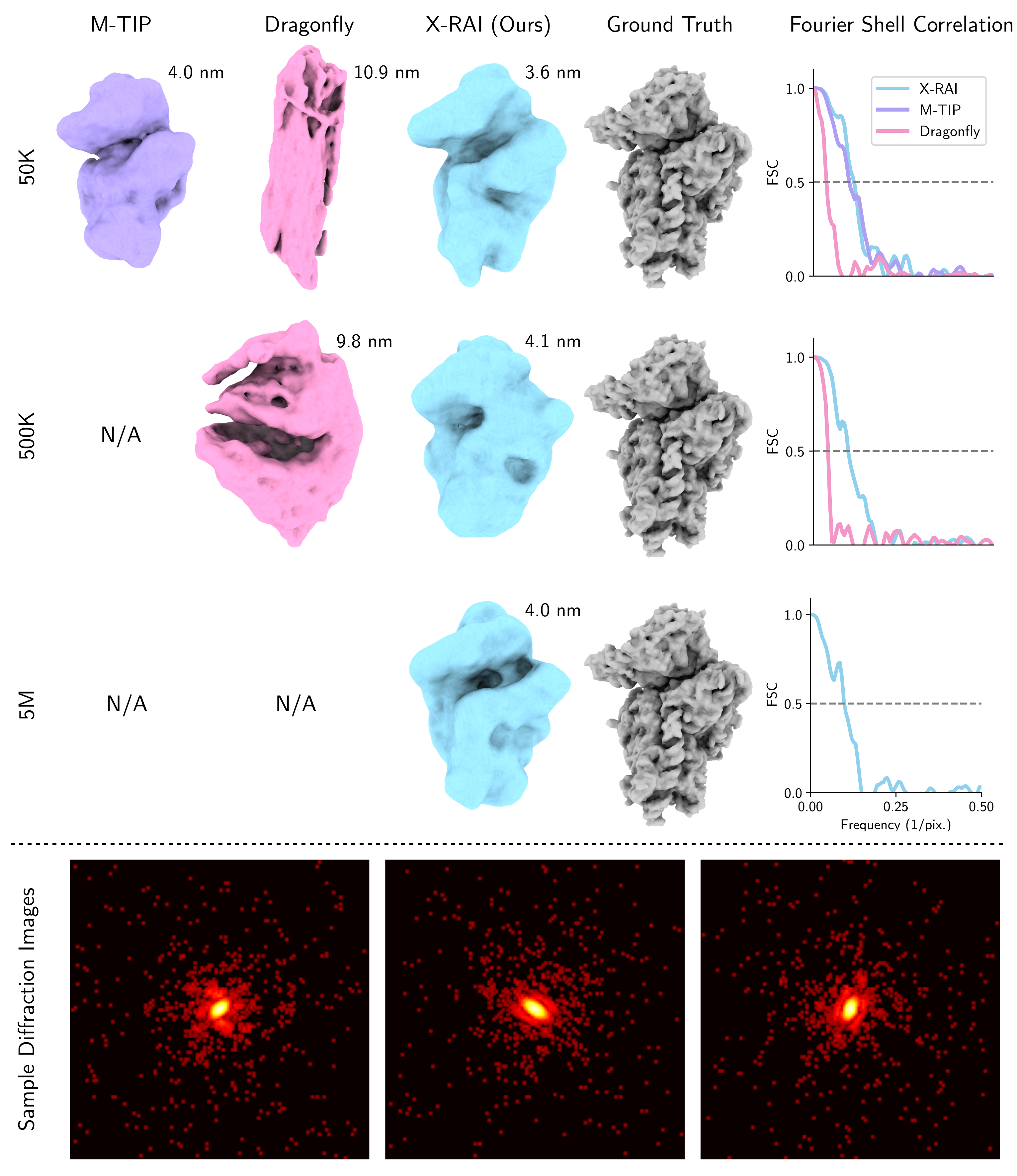}
	\caption{Comparison of X-RAI, M-TIP, and Dragonfly on reconstructing the 40S ribosomal subunit (PDB: 7R4X). The label ``N/A" for M-TIP or Dragonfly indicates failure to process the respective datasets in the allotted time of 48 hours. Reconstructions produced by Dragonfly are scaled down in size in order to fit into the figure.
	}
	\label{fig:supp:7R4X_reconstructions}
\end{figure}

\begin{figure}[ht]
	\centering
	\includegraphics[width=\columnwidth]{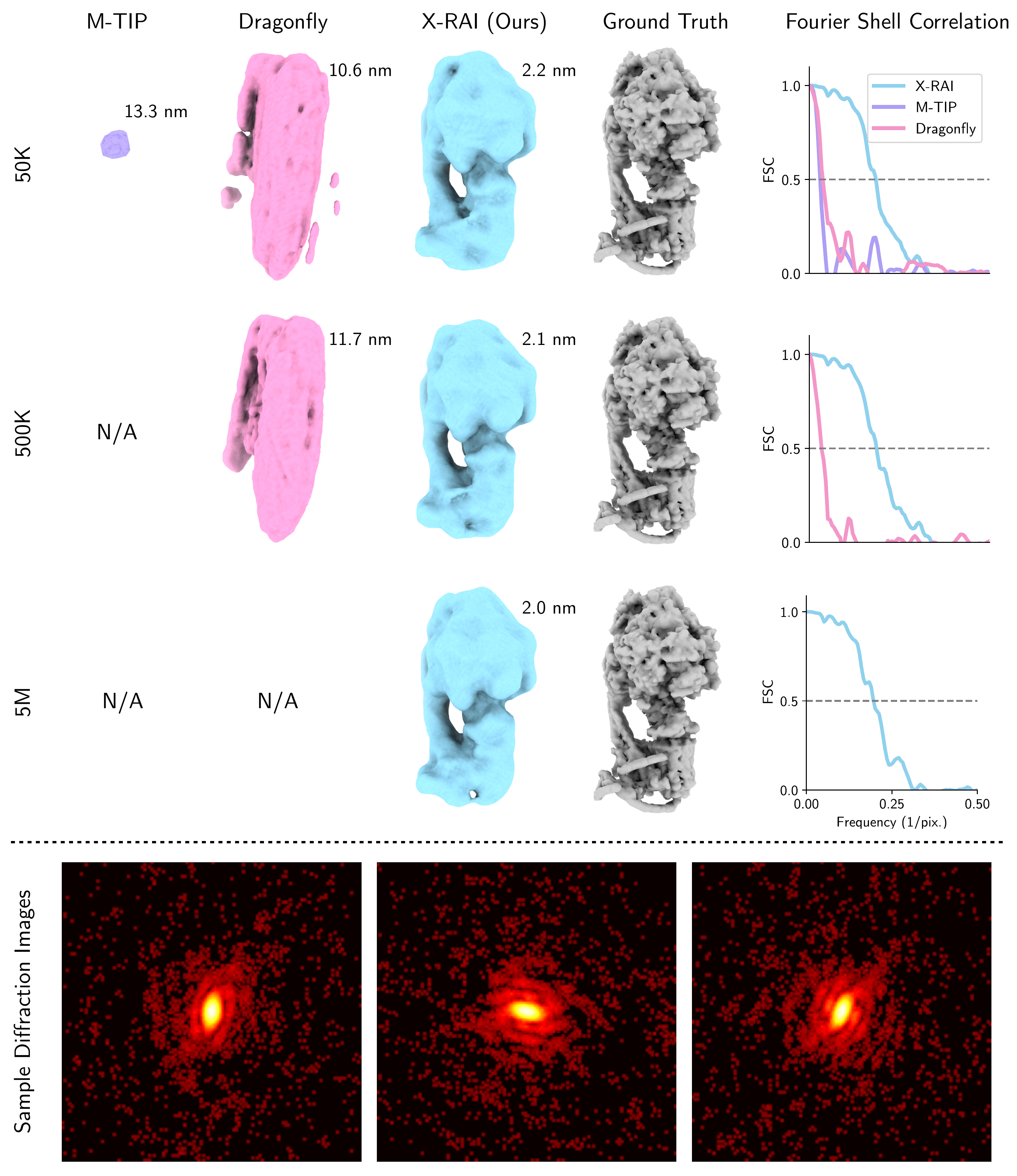}
	\caption{Comparison of X-RAI, M-TIP, and Dragonfly on reconstructing the ATP synthase molecule (PDB: 6J5I). The label ``N/A" for M-TIP or Dragonfly indicates failure to process the respective datasets in the allotted time of 48 hours. Reconstructions produced by Dragonfly are scaled down in size in order to fit into the figure.
	}
	\label{fig:supp:6J5I_reconstructions}
\end{figure}

\begin{figure}[ht]
	\centering
	\includegraphics[width=\columnwidth]{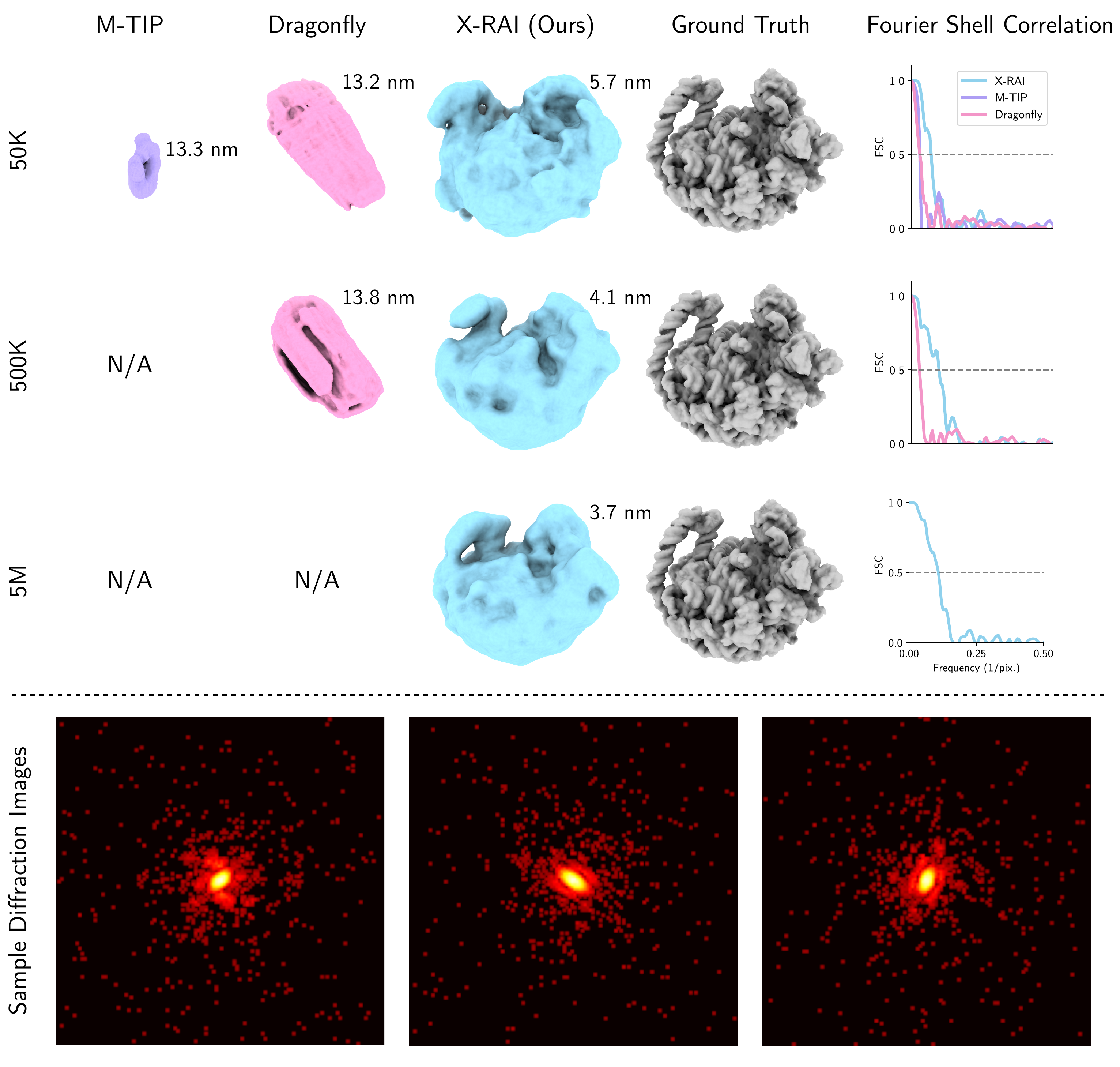}
	\caption{Comparison of X-RAI, M-TIP, and Dragonfly on reconstructing the 50S ribosomal subunit (PDB: 5O60). The label ``N/A" for M-TIP or Dragonfly indicates failure to process the respective datasets in the allotted time of 48 hours. Reconstructions produced by Dragonfly are scaled down in size and re-oriented in order to fit into the figure.
	}
	\label{fig:supp:5O60_reconstructions}
\end{figure}

We present comprehensive reconstruction results on all of the synthetic datasets shown in Table 1 of the main paper. M-TIP fails to reconstruct datasets containing 500 thousand or 5 million images within the allotted time of 48 hours, while Dragonfly times out when processing 5 million images. The results are shown in figures~\ref{fig:supp:7R4X_reconstructions},~\ref{fig:supp:6J5I_reconstructions}, and~\ref{fig:supp:5O60_reconstructions}.

\newpage
\clearpage

\renewcommand{\refname}{Supplementary References}
\printbibliography